%% file: main.tex
\documentclass[sigconf]{acmart}

\usepackage{todonotes}
\usepackage{algorithm}
\usepackage{algpseudocode}
\usepackage[most]{tcolorbox}
\usepackage{multirow}
\usepackage{amssymb}
\usepackage{tabularx}
\usepackage{listings}
\usepackage{fontawesome5}
\usepackage{booktabs}
\usepackage{array}
\usepackage[utf8]{inputenc}
\newcommand{\redacted}[1]{\textbf{REDACTED}}

\AtBeginDocument{%
  }

\setcopyright{acmlicensed}
\copyrightyear{2026}
\acmYear{2026}
\acmDOI{XXXXXXX.XXXXXXX}
\acmConference[32nd SIGKDD Conference on Knowledge Discovery and Data Mining]{Make sure to enter the correct
  conference title from your rights confirmation email}{August 09--13,
  2026}{Jeju, Korea}
\acmISBN{978-1-4503-XXXX-X/2018/06}

\begin{document}


\title[Contextual Bandits and LLMs for Personalized PA]{Tailored Behavior-Change Messaging for Physical Activity: Integrating Contextual Bandits and Large Language Models}

\author{Haochen Song}
\authornote{Both authors contributed equally to this work and share co-first authorship.}
\email{fred.song@mail.utoronto.ca}
\orcid{0009-0006-7825-4993}
\affiliation{%
  \institution{University of Toronto}
  \city{Toronto}
  \state{Ontario}
  \country{Canada}}

\author{Dominik Hofer}
\authornotemark[1]
\email{dominik.Hofer@dhp.lbg.ac.at}
\affiliation{%
  \institution{Ludwig Boltzmann Institute for Digital Health and Prevention}
  \city{Salzburg}
  \country{Austria}
}

\author{Rania Islambouli}
\affiliation{%
  \institution{Ludwig Boltzmann Institute for Digital Health and Prevention}
  \city{Salzburg}
  \country{Austria}
}

\author{Laura Hawkins}
\affiliation{%
  \institution{University of Toronto}
  \city{Toronto}
  \state{Ontario}
  \country{Canada}}

\author{Ananya Bhattacharjee}
\affiliation{%
  \institution{Stanford University}
  \city{Stanford}
  \state{California}
  \country{United States of America}}

\author{Zahra Hassanzadeh}
\affiliation{%
  \institution{University of Toronto}
  \city{Toronto}
  \state{Ontario}
  \country{Canada}}

\author{Jan Smeddinck}
\affiliation{%
  \institution{Ludwig Boltzmann Institute for Digital Health and Prevention}
  \city{Salzburg}
  \country{Austria}
}

\author{Meredith Franklin}
\affiliation{%
  \institution{University of Toronto}
  \city{Toronto}
  \state{Ontario}
  \country{Canada}}

\author{Joseph Jay Williams}
\affiliation{%
  \institution{University of Toronto}
  \city{Toronto}
  \state{Ontario}
  \country{Canada}}

\renewcommand{\shortauthors}{Song, Hofer et al.}

\begin{abstract}
    \input{sections/0_abstract}
\end{abstract}






\maketitle

\input{sections/1_introduction}
\input{sections/2_related-work}
\input{sections/3_exp-models}
\input{sections/4_method}
\input{sections/5_stats-model}
\input{sections/6_results}
\input{sections/7_discussion}
\input{sections/8_conclusion}
% \input{sections/10_acknowledgement}
\bibliographystyle{ACM-Reference-Format}

\bibliography{main}

\newpage
\input{sections/9_appendix}


\end{document}

%% file: sections/0_abstract.tex
Contextual multi-armed bandit (cMAB) algorithms offer a promising framework for adapting behavioral interventions to individuals over time. However, cMABs often require large samples to learn effectively and typically rely on a finite pre-set of fixed message templates. In this paper, we present a hybrid cMABxLLM approach in which the cMAB selects an intervention type, and a large language model (LLM) which personalizes the message content within the selected type. We deployed this approach in a 30-day physical-activity intervention, comparing four behavioral change intervention types: behavioral self-monitoring, gain-framing, loss-framing, and social comparison, delivered as daily motivational messages to support motivation and achieve a daily step count. Message content is personalized using dynamic contextual factors, including daily fluctuations in self-efficacy, social influence, and regulatory focus. Over the trial, participants received daily messages assigned by one of five models: equal randomization (RCT), cMAB only, LLM only, LLM with interaction history, or cMABxLLM. Outcomes include motivation towards physical activity and message usefulness, assessed via ecological momentary assessments (EMAs). We evaluate and compare the five delivery models using pre-specified statistical analyses that account for repeated measures and time trends. We find that the cMABxLLM approach retains the perceived acceptance of LLM-generated messages, while reducing token usage and providing an explicit, reproducible decision rule for intervention selection. This hybrid approach also avoids the skew in intervention delivery by improving support for under-delivered intervention types. More broadly, our approach provides a deployable template for combining Bayesian adaptive experimentation with generative models in a way that supports both personalization and interpretability.

%% file: sections/1_introduction.tex
\section{Introduction}

Physical activity (PA) plays a crucial role in our health and well-being \cite{WHOPhysicalActivity2024}. Growing evidence links insufficient PA to an increased risk of chronic diseases~\cite{anderson2019physical, thornton2022miracle, singh2023effectiveness} with an estimated four to five million annual deaths due to sedentary lifestyles~\cite{lee2012effect, strain2020use}. In response, a wide array of digital health interventions and activity-promoting applications have emerged~ \cite{jorke2025gptcoach, aguilera2024effectiveness, singh2023effectiveness}, designed to support sustained behavior change and encourage regular PA across diverse populations.

The effectiveness of such interventions often depends on a user's shifting daily context and psychological state \cite{haag_bridging_2025, bhattacharjee_investigating_2023}. To address this, researchers have increasingly adopted the Just-In-Time Adaptive Intervention (JITAI) framework, which delivers supportive messages tailored to users' immediate context and individual needs \cite{hardeman_systematic_2019, van_genugten_beyond_2025, jorke2025bloom}. Longitudinal studies have shown that such context-aware approaches can help maintain healthy behaviors over time \cite{shameli_how_2017, feil_systematic_2021}. Yet, real-time fluctuations in mood, stress, self-efficacy, and social influence continue to moderate intervention effectiveness \cite{bhattacharjee_investigating_2023}, presenting a key challenge for adaptive personalization. In practice, this requires to choose which intervention type to deliver and crafting how it is communicated in the moment. 

Building on early work exploring the viability of using adaptive experiments to deploy JITAIs, cMABs provide a structured and interpretable decision-making framework. They enable dynamic selection among a finite set of predefined intervention types based on contextual inputs, such as mood or perceived self-efficacy~\cite{ameko_offline_2020}. However, cMABs often require large sample sizes to learn effectively, and adding more model complexity (e.g., additional psychological or environmental variables) can reduce interpretability and limit performance in practice~\cite{lei_actor-critic_2022, varatharajah_contextual-bandit-based_2022}. In addition, approaches that rely on fixed message templates may struggle to adapt message tone and content to users’ day-to-day context.

Meanwhile, LLMs offer a complementary capability to cMABs by offering flexible intervention types through linguistic personalization. They can adapt the tone, wording, and framing of intervention messages based on user context, generating more relatable and engaging communication \cite{haag_last_2025}. Although LLMs are powerful in language generation, their outputs lack consistency and their internal decision-making is often obscure \cite{bhattacharjee_perfectly_2025, plunkett_self-interpretability_2025}. 

Taken together, cMABs and LLMs motivate a hybrid design that separates decision-making from language generation: cMABs transparently adapt which intervention type to deliver, and LLMs to personalize how that intervention is experessed in natural language. We find this combined approach preserves the interpretability and adaptability of bandit-based selection while improving personalization through flexible message generation, yielding interventions that are more psychologically attuned and behaviorally effective for promoting PA. 

We address two core \textbf{Research Questions} (RQs):

\begin{itemize}
    \item \textbf{RQ1}: To what extent does incorporating contextual factors in the combined cMABxLLM approach influence users' acceptance of intervention messages, compared with other assignment strategies?
    \item \textbf{RQ2}: How do users' daily acceptance ratings and self-reported message evaluations (e.g., relevance, motivational appeal, clarity) relate to longer-term changes in motivation to engage in physical activity (PA)?
\end{itemize}

Our study yields three key takeaways. First, LLM-based personalization produced substantially higher message-acceptance ratings than non-personalized baselines, and the hybrid cMABxLLM achieved acceptance comparable to pure LLM approach while preserving transparent and clear decision logic by separating intervention-type selection from language generation. Second, we observed reliable differences across intervention types: positively framed messages (emphasizing benefits of PA) were rated higher on average, whereas negatively framed messages (emphasizing costs of inactivity) were rated lower, highlighting that \emph{which} intervention type is selected remains important even when language is personalized.Thirdly, evidence for longer-term motivation change was limited in this study: only a subset of participants completed matched pre-post surveys, and estimated changes were small and noisy, motivating larger studies and/or higher-frequency behavioral outcomes (e.g., daily motivation evaluations) to better capture longitudinal effects.

% The following sections describe our theoretical foundation, study design, statistical modeling approach, outcomes, and implications for advancing the hybrid approach.

%% file: sections/2_related-work.tex
\section{Related Work}

We position our contribution in a key design gap: existing work typically adapts either which intervention to deliver (e.g., bandits) or how to express it (e.g., LLMs), whereas our cMABxLLM framework explicitly separates intervention-type selection from language generation to enable scalable, interpretable JITAIs for physical activity.

\subsection{Contextual Bandits for Physical Activity Interventions}

Multi-armed bandit (MAB) and Contextual Multi-Armed Bandits (cMABs) algorithms have been widely applied to tailor PA interventions for sedentary adults by responding to real-time contextual cues, such as location, time of day, and recent movement. For instance, the HeartSteps trial by Klasnja et al. demonstrated that context-sensitive prompts produced immediate increases in step counts; however, these gains diminished over time, highlighting challenges around habituation and intervention fatigue \cite{klasnja_efficacy_2019}. Other research has reported more sustained benefits: the MyBehavior app, led to users walking an extra ten minutes each day \cite{rabbi_automated_2015}. Recent reviews also highlight MABs and cMABs as a promising approach for real-time personalization in digital health, emphasizing adaptive intervention delivery as a pathway to improved engagement and behavior change \cite{svihrova_designing_2025}. More recent work has studied conversational contextual bandits, where a system interacts over multiple turns to elicit key information before acting, illustrating that bandit personalization can be sequential and intereactive~\cite{Liu_TermSelection_2025}.

Together, these findings underscore the potential of cMABs for learning which intervention to deliver under changing contexts to encourage PA. However, most bandit-based systems work over a small, fixed set of predefined templates or strategies, limiting their ability to adapt message phrasing and tone (which have been previously demonstrated to influence perceived relevance, acceptance and long-term engagement~\cite{Hekler_tailored_messages_feasibility_2023, Wu_AIM_2024}.

\subsection{LLMs for Motivational Message Generation}

Parallel work has explored the use of large language models (LLMs) to generate motivational and theory-grounded messages. Jörke et al. employed GPT-4 with prompt chaining to emulate motivational interviewing during a sixteen-participant technology probe; it upheld a supportive tone in 84 percent of its utterances and favored open-ended questions over unsolicited advice \cite{jorke2025gptcoach}. Other studies have linked LLM with behavioral framework such as COM-B to tailor responses to users' motivational states, with expert raters evaluating these outputs as more empathetic and actionable in simulated dialogues \cite{hegde_infusing_2024}. In a JITAI study conducted by Willms and Liu, ChatGPT rapidly produced thirteen theory-based lesson messages for parents to promote child activity, demonstrating the model's capacity for quick, customizable content creation (although expert review was still necessary) \cite{willms_exploring_2024}. Haag et al. tested GPT-4 for both triggering decisions and message generation in a cardiac rehabilitation JITAI context, finding that LLM-generated interventions outperformed those created by both laypersons and healthcare professionals across metrics of appropriateness, engagement, effectiveness, and professionalism, although the evaluation relied on persona-based scenarios rather than real-world deployment \cite{haag_last_2025}. All together, research in this domain has demonstrated LLMs provide benefits such as personalized, supportive communication and fast content generation. Notably, these studies face limitations such as their brief durations, small sample sizes, absence of real-world behavior outcomes, and need for expert oversight.

% Other works on conversational and sequential personalization research shows that personalization is often achieved through multi-turn interaction, where system adapt responses as it learns user preferences~\cite{Jeon_VisionLLM_2025}. At the same time, recent work questions whether LLMs can reliably capture sequential structure in recommendation setting, showing the risk of relying on LLMs alone for sequential decision making without any explicit and auditable policy~\cite{Kim_sequence_2025}. 

% Overall, while LLMs can make reasonable next-step choices, using them as the primary decision-maker can be computationally costly and introduces reliability and reproducibility risks such as sensitivity to the choice of prompts and model versions. These challenges motivate explorations into separating decision-making from language generation.

\subsection{Hybrid cMABs-LLM Approaches for Adaptive Interventions}
Recent investigations have begun to integrate hybrid AI systems that pair decision making algorithms with LLMs to create more adaptive health-promotion strategies. Karine and Marlin (2025) introduced a hybrid framework in which a Thompson Sampling bandit suggests intervention options, and an LLM filters these suggestions based on user-stated preferences, leading to improved personalization in a PA setting \cite{karine_combining_2025}. Alamdari et al. (2024) proposed using LLM-generated synthetic data to warm-start CMABs, thereby reducing initial regret and accelerating the learning process \cite{alamdari_jump_2024}. In another study, Brooks et al. (2024) conceptualized LLMs as stochastic treatment generators and developed GAMBITTS-a Thompson Sampling variant that accounts for uncertainty in the content generated by the model-resulting in lower regret in simulated PA interventions \cite{brooks_generator-mediated_2025}. Wang et al. (2025) combined a context-aware bandit with an LLM-driven chatbot to deliver tailored motivational messages in the realm of sleep health, demonstrating enhanced engagement and improved health outcomes compared to non-adaptive baselines \cite{wang_exploring_2025}. Other research also explores hybrid architectures that pair LLM components with sequential recommendation models, reflecting a broader trend toward separating language generation from the core decision mechanism~\cite{Lv_LLMrecommendation_2025}.
Together, these efforts suggest that coupling cMABs with LLMs can produce interventions that are more responsive and personalized, with potential gains in behavior-change effectiveness. At the same time, they surface open feasibility and system-design questions for real-world deployment.

%% file: sections/3_exp-models.tex
\section{Experiment Delivery Models}
\label{sec:exp_models}

This study cross-compares five experimental models. Across all models, we collect the same participant inputs and outcomes so that participants are blinded to their experimental condition assignment; models differ only in how the intervention type is selected and how the message is generated. 

Let $X_t$ denote the context observed at each decision time $t$, including individual factors such as self-efficacy, social influence, and regulatory focus. We denote the intervention (arm) space by $\mathcal{A}$, where each arm corresponds to an intervention type and $|\mathcal{A}|=4$. We additionally collect participants' free-text written narrative input $\mathcal{L}_t$ at each time step. At time $t$, the assignment policy selects an arm $A_t \in \mathcal{A}$ based on the current context $X_t$.

After delivering an intervention, we observe an outcome (reward) $R_t$. Let $D_t$ denote the accumulated information history up to time $t$, with $D_0=\emptyset$ and, for $t\ge 1$, 
\[
D_t = D_{t-1}\cup\{(X_{t-1},A_{t-1},R_{t-1})\},
\]
where each $D_t$ contains the observed context, chosen arm, and realized reward.

\subsection{Randomized Controlled Trial (RCT)}
\label{sec:rct}

In the \textbf{RCT} condition, the intervention arm $A_t \in \mathcal{A}$ is assigned uniformly at random at each decision time step $t$, independent of the participant's context $X_t$, free-text input $\mathcal{L}_t$, and historical dataset $D_t$. All other study procedures (data collection, message delivery, and outcome measurement) are held constant across conditions, so that any differences in outcomes are attributable to the assignment mechanism rather than differences in participant experience.

\subsection{Contextual Multi-Armed Bandit (cMAB)}
\label{sec:cmab}

CMABs (Algorithm \ref{alg:cMAB}) dynamically select interventions based on contexts $X_t$ at each time step $t$. Following the Bayesian bandit tradition, we model arm rewards with a Bayesian linear contextual model and use posterior sampling (i.e., Thompson Sampling) to balance exploration and exploitation~\cite{chu_contextual_2011, thompson_likelihood_1933}. Specifically, conditional on context $X_t$ and arm $a$, the expected reward is
\[
\mu_{a,t}=X_t^\top \beta + D(a)^\top \gamma,
\]
where $\beta$ is a shared context coefficient vector and $\gamma$ captures arm-specific main effects (with $D(a)$ the one-hot encoding of arm $a$). We maintain a posterior distribution over parameters,
\[
\begin{bmatrix}\beta\\\gamma\end{bmatrix}\mid D_t \sim \mathcal{MVN}(m_t,S_t),
\]
with posterior mean $m_t$ and covariance $S_t$, inducing a posterior over each arm's expected reward. At each time step, Thompson Sampling draws $\bigl[\beta^{(t)},\gamma^{(t)}\bigr]$ from this posterior and selects the arm with the largest sampled expected reward; after observing $R_t$, we update $(m_t,S_t)$ using $D_{t+1}$. 

\begin{algorithm}[t]
\caption{Contextual Thompson Sampling with Linear Payoff}
\label{alg:cMAB}
\begin{algorithmic}[1]
\For{$t = 1,2,\dots$}
    \State Observe context $X_t$
    \State Sample $\theta^{(t)}=\begin{bmatrix}\beta^{(t)}\\ \gamma^{(t)}\end{bmatrix} \sim \mathcal{MVN}(m_t,S_t)$
    \State Choose $A_t=\arg\max_{a\in\mathcal{A}}\left(X_t^\top \beta^{(t)} + D(a)^\top \gamma^{(t)}\right)$
    \State Deliver intervention $A_t$ and observe reward $R_t$
    \State Update $D_{t+1}=D_t \cup \{(X_t,A_t,R_t)\}$ and posterior $(m_{t+1},S_{t+1})$
\EndFor
\end{algorithmic}
\end{algorithm}

\subsection{Large Language Models (LLM-only)}

In use of the LLM-only model, (Algorithm \ref{algorithm2}) we hold and use the same structured context \( X_t \) as before, but introduce the user-generated free-text input \( \mathcal{L}_t \) to the model, as shown in Algorithm~\ref{algorithm2}. This response captures the user's situational context and personal narrative, offering a closer alignment with users' linguistically sophisticated content. The full prompt template is provided in Appendix~\ref{appx:llm_only_template}.

At each time step, the LLM first selects an intervention type \( A_t \in \mathcal{A} \) from the \textbf{finite} set of possible intervention types. It then generates a customized intervention message \( M_t \in \mathcal{M} \), where \( \mathcal{M} \) is an \textbf{infinite} set of linguistically adaptive messages derived from the base arm space \( \mathcal{A} \). This process is modeled as:
\[
M_t \sim f_{\text{LLM}}(X_t, \mathcal{L}_t \mid A_t).
\]
Here, \( f_{\text{LLM}} \) is treated as an unobserved (black-box) mapping, instantiated by GPT-4 Turbo, that produces a tailored message based on the structured context \( X_t \), the user's free-text input \( \mathcal{L}_t \), and the selected base intervention \( A_t \). Unlike selecting a fixed message from \( \mathcal{A} \), the LLM adapts the intervention's language and framing to the user's current psychological and situational context, though the specific decision-making process is not directly observable.

\begin{algorithm}[t]
\caption{LLM-based Personalized Messaging (LLM-only)}
\label{algorithm2}
\begin{algorithmic}[1]
\For{$t = 1,2,\dots$}
    \State Observe context $X_t$ and user-written input $\mathcal{L}_t$
    \State LLM selects intervention type $A_t \in \mathcal{A}$ given $(X_t,\mathcal{L}_t)$
    \State Generate message $M_t = f_{\text{LLM}}(X_t,\mathcal{L}_t \mid A_t)$
    \State Deliver $M_t$ and observe reward $R_t$
    \State Update $D_{t+1}=D_t \cup \{(X_t,A_t,R_t)\}$ \footnotemark
\EndFor
\end{algorithmic}
\end{algorithm}

\footnotetext{We log $(X_t, A_t, R_t)$ into $D_t$ for evaluation, but $D_t$ is not provided to the LLM in the LLM-only condition.}

\subsection{Large Language Models with History Tracing (LLM-tracing)}
\label{sec:llm_tracing}

The \textbf{LLM-tracing} model (Algorithm~\ref{algorithm2_1}) extends the LLM-only approach by incorporating each participant's recent interaction history into the prompt. Specifically, we construct a history trace $H_t$\footnotemark from the most recent 10 days of interactions, which includes prior contexts, selected intervention types, observed rewards, and previously generated messages. This enables message generation to reflect longitudinal patterns (e.g., repeated barriers, sustained motivation, or recent disengagement). We use the same structured context $X_t$ and free-text input $\mathcal{L}_t$ as in the LLM-only condition, augmented with $H_t$. The full prompt template is provided in Appendix~\ref{appx:llm_tracing_template}.

\footnotetext{Concretely, we define $H_t=\{(X_{t-i},A_{t-i},R_{t-i},M_{t-i})\}_{i=1}^{10}$, i.e., the last 10 interactions including the corresponding messages $M_{t-1},\ldots,M_{t-10}$.}

At each time step, the LLM-tracing model (instantiated with GPT-4 Turbo) selects a base intervention $A_t \in \mathcal{A}$ and generates a personalized message $M_t$, conditioned on the augmented input $(X_t,\mathcal{L}_t,H_t)$:
\[
M_t = f_{\text{LLM-tracing}}(X_t,\mathcal{L}_t,H_t \mid A_t).
\]

\begin{algorithm}[t]
\caption{LLM-tracing Personalized Messaging}
\label{algorithm2_1}
\begin{algorithmic}[1]
\For{$t = 1,2,\dots$}
    \State Observe context $X_t$ and user-written input $\mathcal{L}_t$
    \State Construct history trace 
    \[
    H_t \leftarrow \Big(\{(X_{t-i},A_{t-i},R_{t-i})\}_{i=1}^{10},\ \{M_{t-i}\}_{i=1}^{10}\Big)
    \]
    \State LLM selects intervention type $A_t \in \mathcal{A}$ given $(X_t,\mathcal{L}_t,H_t)$
    \State Generate message $M_t = f_{\text{LLM-trace}}(X_t,\mathcal{L}_t,H_t \mid A_t)$
    \State Deliver $M_t$ and observe reward $R_t$
    \State Update $D_{t+1}=D_t \cup \{(X_t,A_t,R_t)\}$
\EndFor
\end{algorithmic}
\end{algorithm}

\subsection{Hybrid Approach Using cMABs and LLMs (cMABxLLM)}

In the hybrid cMABxLLM approach (Algorithm~\ref{algorithm3}), we separate \emph{intervention-type selection} from \emph{message generation}. The intervention type $A_t$ is selected using the same contextual Thompson sampling procedure as in Section~\ref{sec:cmab}. Concretely, at each time step $t$, we draw
\[
\begin{bmatrix}\beta^{(t)}\\ \gamma^{(t)}\end{bmatrix} \sim \mathcal{MVN}(m_t,S_t),
\]
and choose the arm with the largest sampled expected reward,
\[
A_t=\arg\max_{a\in\mathcal{A}}\left(X_t^\top \beta^{(t)} + D(a)^\top \gamma^{(t)}\right).
\]
After the intervention type $A_t$ is chosen, the LLM generates the message content conditioned on $A_t$ using the current context $X_t$ and the participant's written input $\mathcal{L}_t$:
\[
M_t = f_{\text{LLM}}(X_t,\mathcal{L}_t \mid A_t). \footnotemark
\]

\footnotetext{The full prompt template is provided in Appendix~\ref{appx:cmabxllm_template}.}

\begin{algorithm}[t]
\caption{cMABxLLM Combined Assignment}
\label{algorithm3}
\begin{algorithmic}[1]
\For{$t = 1,2,\dots$}
    \State Observe context $X_t$ and user-written input $\mathcal{L}_t$
    \State Sample $\begin{bmatrix}\beta^{(t)}\\ \gamma^{(t)}\end{bmatrix} \sim \mathcal{MVN}(m_t,S_t)$
    \State Choose intervention type
    \[
    A_t=\arg\max_{a\in\mathcal{A}}\left(X_t^\top \beta^{(t)} + D(a)^\top \gamma^{(t)}\right)
    \]
    \State Generate message content: $M_t = f_{\text{LLM}}(X_t,\mathcal{L}_t \mid A_t)$
    \State Deliver $M_t$ and observe reward $R_t$
    \State Update $D_{t+1}=D_t \cup \{(X_t,A_t,R_t)\}$ and posterior $(m_{t+1},S_{t+1})$
\EndFor
\end{algorithmic}
\end{algorithm}

%% file: sections/4_method.tex
\section{Methodology}
In this section, we describe the study protocol, intervention message design, outcome collection, and LLM prompting details used to evaluate the experimental models in Section~\ref{sec:exp_models}, including how rewards are logged and (when applicable) used to update the models.

\subsection{Study Design}
% The research protocol was approved by \redacted{the University of Toronto (\#protocol number} Research Ethics Board. 

The study design follows a 30-day protocol: days 1 and 30 are reserved for pre- and post-study evaluation, a 5-minute mid-study interview on the 15th and 16th day, and an optional post-study interview was offered on the 30th day. The other days 2 - 29 (including days where the mid-study interviews took place) were dedicated to daily intervention delivery and assessment. On day 1, participants complete a pre-study questionnaire that includes the Behavioral Regulation in Exercise Questionnaire (BREQ-3) to assess motivation based on Self-Determination Theory (SDT)~\cite{cid_behavioral_2018}; Perceived Trust in Automation using the PAICE scale \cite{scantamburlo_artificial_2023}; and socio-demographic information including gender, education level, and digital skill proficiency. On day 30, participants complete a post-study survey focused on their experience  with the intervention messages including perceptions of personalization,  appropriateness, and comparative evaluation across intervention types.

During the 28-day intervention phase (days 2-29), 93 participants enrolled and $n=54$ actively participated by completing a brief daily ecological momentary assessment (EMA) survey ($\sim$5 min) throughout the study. This daily EMA collected participants' context ($X_t$) and free-text input ($\mathcal{L}_t$). After completing the EMA each day, participants were assigned (via micro-randomization) to one of five experimental models described in Section~\ref{sec:exp_models}: \textbf{RCT}, \textbf{LLM-only}, \textbf{cMAB-only}, \textbf{LLM-tracing}, or the \textbf{combined cMABxLLM} model. Micro-randomization was repeated independently each day, enabling within-subject comparisons across models.

\begin{itemize}
    \item \textbf{Mood (1-100):} Self-reported current mood (higher = better mood).
    \item \textbf{Stress Level (1-100):} Perceived stress / difficulty managing upcoming tasks (higher = more stressed).
    \item \textbf{Self-Efficacy (1-100):} Confidence in engaging in planned PA (higher = stronger belief in meeting PA goals).
    \item \textbf{Social Influence (1-100):} Likelihood of joining others in PA if prompted (higher = more likely).
    \item \textbf{Regulatory Focus ($-3$ to $+3$):} Prevention vs.\ promotion orientation (negative = prevention/loss-avoidance; positive = promotion/growth/achievement).
    \item \textbf{Written Narrative:} A brief description of daily events and their impact on mood, stress, or PA.
\end{itemize}

Although mood and stress are collected daily, they are not used in the real-time cMAB or LLM assignment procedures. Instead, they are included in  post-hoc analyses to assess any potential confounding or moderating effects of high-frequency contextual factors (those that vary hourly) versus lower-frequency ones (those that change daily). To clarify the variables and written narrative responses, Table~\ref{tab:variables_used} provides an overview of which contextual variables are used for model training in each experimental group.

\renewcommand{\arraystretch}{1}
\begin{table}[h]
\centering
\small
\caption{Inputs used by each experimental model for assignment and message generation (\checkmark = used by the model; \texttimes = not used).}
\resizebox{\columnwidth}{!}{%
\begin{tabular}{|p{2.8cm}|c|c|c|c|c|}
\hline
\textbf{Variable}  & \textbf{RCT} & \textbf{cMAB} & \textbf{LLM-only} & \textbf{LLM-tracing} & \textbf{cMABxLLM} \\ \hline
Mood               & \texttimes   & \texttimes    & \texttimes   & \texttimes            & \texttimes        \\ \hline
Stress Level       & \texttimes   & \texttimes    & \texttimes   & \texttimes            & \texttimes        \\ \hline
Self-Efficacy      & \texttimes   & \checkmark    & \checkmark   & \checkmark            & \checkmark        \\ \hline
Social Influence   & \texttimes   & \checkmark    & \checkmark   & \checkmark            & \checkmark        \\ \hline
Regulatory Focus   & \texttimes   & \checkmark    & \checkmark   & \checkmark            & \checkmark        \\ \hline
Written Narrative ($\mathcal{L}_t$) & \texttimes & \texttimes & \checkmark & \checkmark          & \checkmark        \\ \hline
Interaction History ($H_t$) & \texttimes & \texttimes & \texttimes & \checkmark          & \texttimes        \\ \hline
\end{tabular}}

\label{tab:variables_used}
\end{table}

\begin{figure}[h]
    \centering
    \caption{Study timeline (Days 1-30): pre-study questionnaire (Day 1), daily EMA and message delivery (Days 2-29), and post-study questionnaire (Day 30).}
    \includegraphics[width=1.0\linewidth]{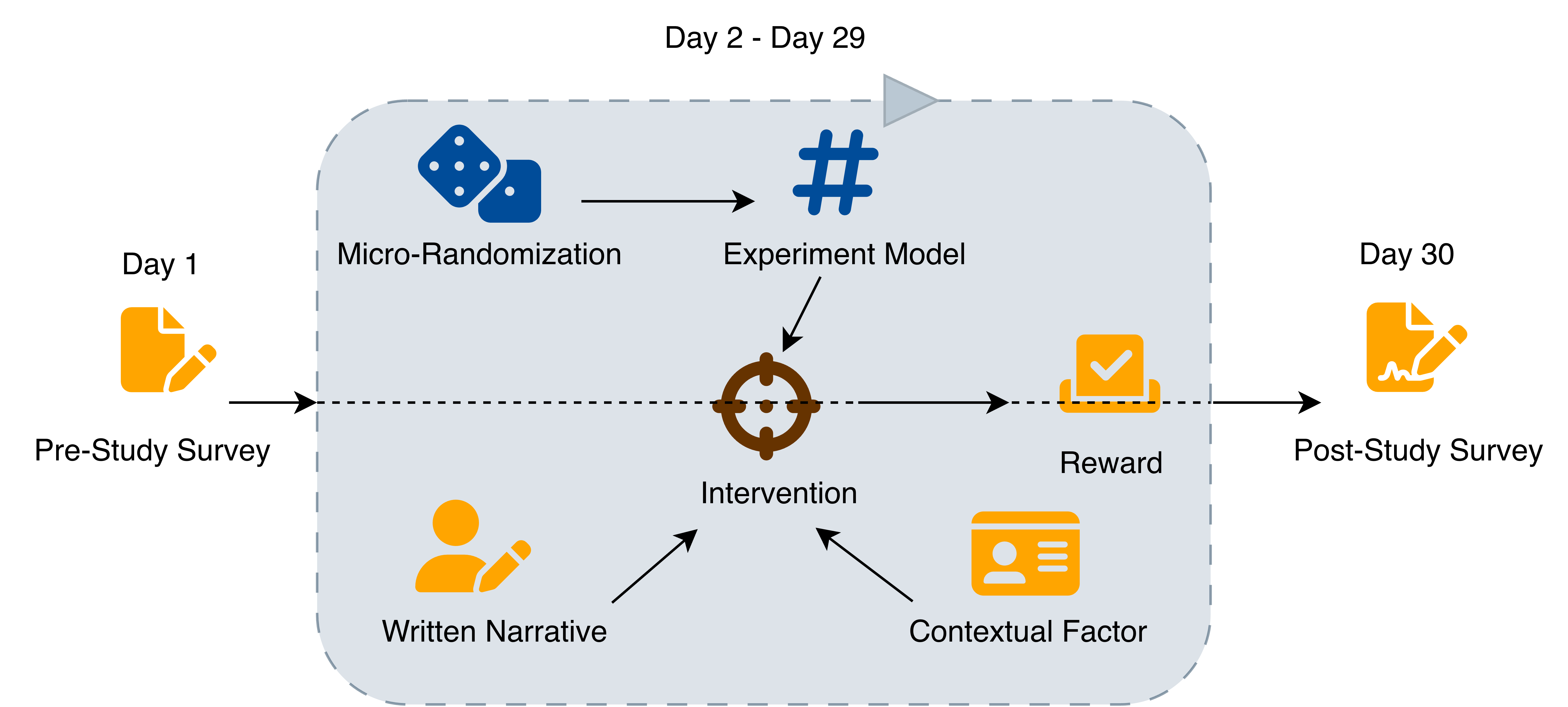}
    \label{fig:study_design}
\end{figure}

Each day, participants receive one motivational message at a participant-selected preferred time, generated by the assigned experimental model. Figure~\ref{fig:study_design} provides an overview of the study timeline and key steps. We describe the main components below.

\subsection{Intervention Messages}
\label{sec:intervention_messages}
Each motivational message corresponds to one of four behavior-change intervention types. In non-LLM conditions, we deliver a predefined template message matched to the assigned intervention type (templates in Appendix~\ref{appx:non_llm_message}); in LLM-based conditions, the LLM generates diverse messages within the same type, conditioned on participant context (examples in Appendix~\ref{appx:llm_message}). 

We selected these four intervention types based on established behavior-change literature and prior evidence in physical-activity messaging: 

\begin{enumerate}
    \item \textbf{behavioral self-monitoring} prompts reflection on recent activity and progress tracking~\cite{michie_effective_2009};
    \item \textbf{gain-framing} emphasizes benefits of exercising, which has shown favorable effects on PA in several studies~\cite{tversky_framing_1981, latimer_systematic_2010};
    \item \textbf{loss-framing} emphasizes costs of inactivity and serves as a theoretically motivated contrast, with mixed evidence relative to gain-framing~\cite{okeefe_relative_2007, gallagher_health_2012, latimer_systematic_2010}; and
    \item \textbf{social comparison} leverages normative information about others' behavior, with emerging evidence for PA promotion~\cite{chapman_goals_2016, nastasi_stepping_2022}
\end{enumerate}

\subsection{Reward and Feedback Collection}

Once viewing their assigned intervention message, participants are directed to a response page where they report two key outcomes: (1) \textbf{message acceptance}-how appropriate, relatable, or useful they found the message; and (2) \textbf{momentary motivation}-their current motivation to engage in PA. We treat message acceptance as the reward signal $R_t$ and log it for analysis in all conditions; however, $R_t$ is used to update the bandit policy only in cMAB-active conditions (cMAB-only and cMABxLLM).

After this, a participants complete a brief feedback survey. This optional qualitative survey invites participants to reflect on the personalization of the message they received, its relevance to their context, and to offer suggestions for improvement. These narrative responses help us explore perceived personalization, but are not used to update the models during the study.

\subsection{Prompt Example Used in the LLM Model}

We provide below Example 1 to illustrate the general prompting workflow used in all LLM-involved conditions (LLM-only, LLM-tracing, and cMABxLLM). To generate personalized intervention messages, we provide the LLM with both structured and unstructured participant information. Each request consists of (i) a \textit{system} prompt that specifies the task and the available intervention message types, and (ii) a \textit{user} prompt containing the participant's current psychological context and written self-reflection. Example messages are provided in Section~\ref{sec:intervention_messages}, and full prompt templates are provided in Appendix~\ref{appx:LLM_template_all}.

\begin{tcolorbox}[
  float,
  floatplacement=t,
  title={\textbf{Example 1: Prompting workflow for LLM-based message generation (GPT-4 Turbo).}},
  label={box:LLM_prompt_example},
  enhanced,
  breakable,
  colback=white,
  colframe=white,
  boxrule=0.1pt,
  coltitle=black,
  arc=0mm,
  left=0mm,right=0mm,top=0mm,bottom=0mm, % tighten outer padding
]

\begin{tcolorbox}[
  colback=blue!5!white,
  colframe=blue!50!black,
  title={\faRobot\ \textbf{System Prompt}},
  enhanced,
  breakable,
  left=1.5mm,right=1.5mm,top=0.8mm,bottom=0.8mm, % tighten inner padding
  boxsep=0.8mm,
  before skip=0mm,
  after skip=1.0mm % small gap to next box
]
You are an intelligent healthcare assistant tasked with generating personalized health intervention messages to help individuals increase their daily step count. There are four types of intervention available, each defined in the study design:
\begin{itemize}\setlength{\itemsep}{0pt}\setlength{\parskip}{0pt}\setlength{\topsep}{2pt}
    \item Behavioral Self-Monitoring + Feedback
    \item Gain-Framed Messaging
    \item Loss-Framed Messaging
    \item Social Norms \& Comparison Feedback
\end{itemize}
For each participant, you will receive contextual information including self-efficacy, regulatory focus, social influence, and a personal reflection. Based on this information, select one appropriate intervention type and personalize the message using the corresponding template provided in the study design.
\end{tcolorbox}

\begin{tcolorbox}[
  colback=yellow!5!white,
  colframe=orange!80!black,
  title={\faUser\ \textbf{User Prompt}},
  enhanced,
  breakable,
  left=1.5mm,right=1.5mm,top=0.8mm,bottom=0.8mm,
  boxsep=0.8mm,
  before skip=0mm,
  after skip=0mm
]
\textbf{Self-efficacy}: 72/100 (higher values indicate greater confidence in maintaining PA) \\
\textbf{Social influence}: 64/100 (higher values indicate greater responsiveness to encouragement from others) \\
\textbf{Regulatory focus}: +3 (positive values indicate gain orientation; negative values indicate loss orientation; range: $-6$ to $+6$) \\
\textbf{Written Narrative}: \textit{``I've been stressed but walking helps clear my mind.''}
\end{tcolorbox}

\end{tcolorbox}

In the LLM-only and LLM-tracing condition, the model receives all four intervention types and is responsible for selecting and customizing the message. In contrast, in the combined cMABxLLM condition, the intervention type is pre-selected by the cMAB and passed to the LLM, which then generates a message tailored to the participant's psychological context and reflection. This design allows us to isolate and evaluate the added value of LLM-based message personalization, both independently and in combination with cMAB-driven intervention selection.
=

%% file: sections/5_stats-model.tex
\section{Statistical Modeling}

\begin{figure}[h]
    \centering
    \caption{Statistical Directed Acyclic Graphs (DAG) describing the modeling structure of our study.}
    \includegraphics[width=0.9\linewidth]{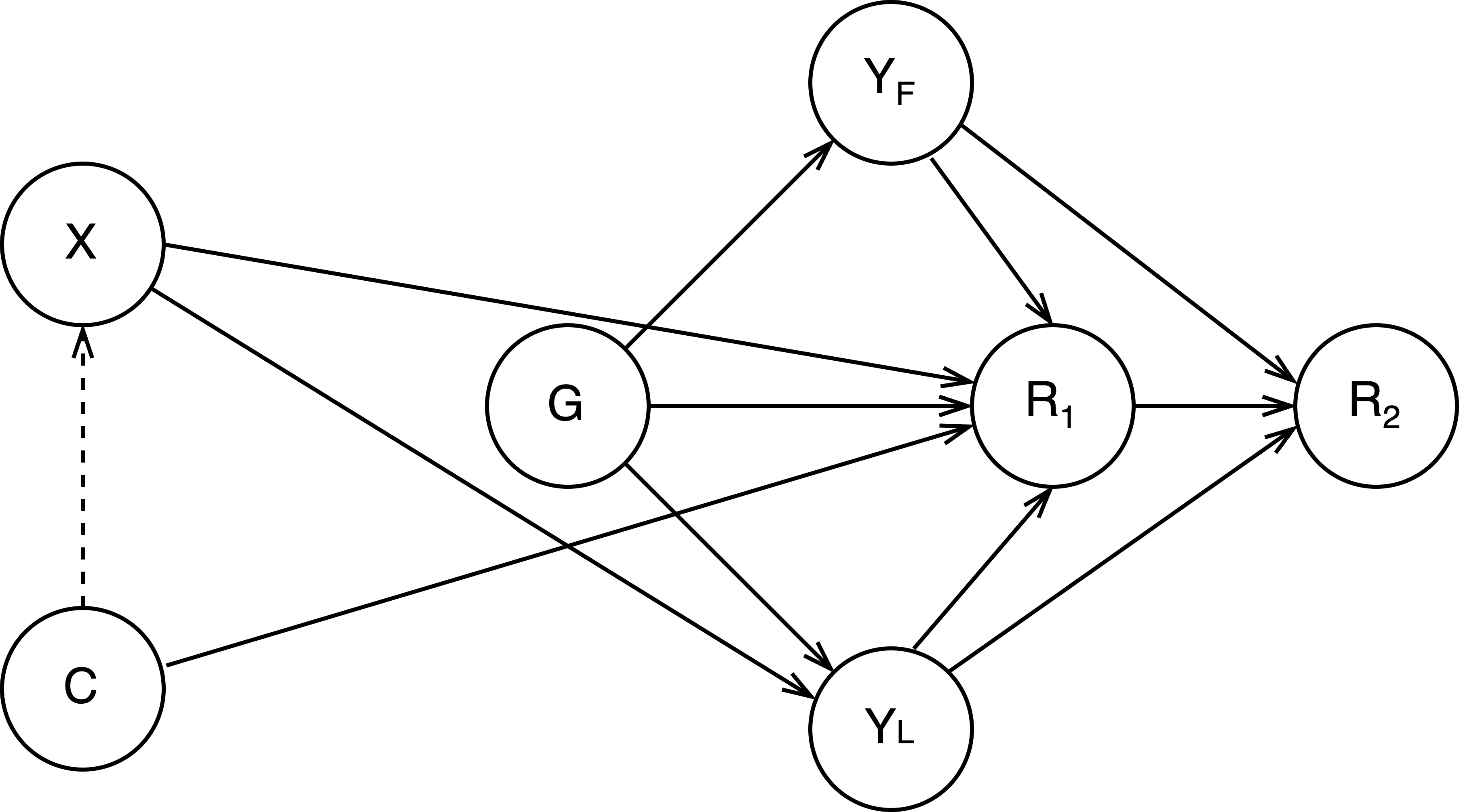}
    \label{fig:dag}
\end{figure}

To address our research questions, we fit two statistical models, one corresponding to each question.
Figure~\ref{fig:dag} summarizes the hypothesized directed acyclic graph (DAG) for our study.
In this diagram, each directed edge indicates an assumed direct dependence (potential causal pathway) between variables.

The experimental condition ($G$) determines which type of intervention message a participant receives: either a fixed template ($Y_F$), as used in the RCT and cMAB models, or a personalized message ($Y_L$)generated by the LLM-only, LLM-tracing, or cMABxLLM models. Contextual features ($X$), such as self-efficacy or social influence, inform the generation of $Y_L$ messages but do not influence ($Y_F$), which are selected from static templates.

Both message types, along with the experimental model $G$ and contextual features $X$, are expected to influence how participants respond to the message, operationalized as message acceptance ($R_1$). In turn, message acceptance, together with context factors, may influence participants' overall motivation to engage in PA ($R_2$). We also include a set of collected variables ($C$) such as mood and stress as potential confounders that may affect $X$ or outcomes indirectly. These potential confounders are included in the study to enable post-hoc adjustment.

\textbf{Variable notations:}
\begin{itemize}
    \item $X$: Contextual features (e.g., self-efficacy, social influence)
    \item $C$: Potential confounders (e.g., mood, stress level)
    \item $G$: Experimental model
    \item $Y_F$: Fixed message content (RCT and cMAB models)
    \item $Y_L$: Personalized message content (LLM and cMABxLLM models)
    \item $R_1$: Message acceptance rating
    \item $R_2$: Change in motivation toward PA (pre- to post-study)
\end{itemize}

\textbf{RQ1} evaluates how different intervention models ($G$) and contextual features ($X$) influence message acceptance ($R_1$). To answer this, we compare $R_1$ outcomes across all four models using offline post-study analysis. The micro-randomization design, in which each participant is randomly assigned to one model per day, mitigates individual-level confounding  by distributing conditions evenly within subjects.

We fit a linear mixed-effects model with fixed effects for the intervention model ($G$) and contextual features ($X$), and random intercepts for participants to account for repeated measures. Message acceptance ($R_1$) is treated as a continuous variable on a 5-point Likert scale from 1 (not acceptable) to 5 (liked very much). We hypothesize that fixed-content models (RCT and cMAB-only) will have lower average acceptance than personalized models (LLM-only, LLM-tracing, and cMABxLLM), although the relative performance of LLM-only, LLM-tracing, and cMABxLLM is not a priori certain. 

\begin{equation}
R_{1}
= \alpha
+ G^\top \boldsymbol{\beta}_{G}
+ A^\top \boldsymbol{\beta}_{A}
+ X^\top \boldsymbol{\beta}_{X}
+ \bigl(A\otimes X\bigr)^\top \boldsymbol{\beta}_{AX}
+ C^\top \boldsymbol{\beta}_{C}
+ b_{0i}
+ \varepsilon,
\label{eq:acceptance_rating}
\end{equation}

where $\boldsymbol{\beta}_{G},\boldsymbol{\beta}_{A},\boldsymbol{\beta}_{X},\boldsymbol{\beta}_{AX},\boldsymbol{\beta}_{C}$ are the fixed-effect coefficient vectors.
$G$ denotes the experimental model condition, $A$ is the intervention type, $X$ is the participant’s contextual state, and $A\otimes X$ denotes their interaction terms.
$C$ includes potential confounders (e.g., mood, stress, and time-in-study terms).
We include a participant-specific random intercept $b_{0i}$ to account for repeated measures within participants, and $\varepsilon$ is the residual error term.
This model captures the pathways $G \to Y_F \ \&\ Y_L \to R_1$ and $C \dashrightarrow X \to R_1$ in the DAG (Figure~\ref{fig:dag}).

\textbf{RQ2} examines the longer-term (longitudinal) effects of intervention on participants' motivation to engage in PA. Motivation change ($R_2$) is measured as the difference between post-study and pre-study measures using the BREQ-3 survey \cite{cid_behavioral_2018} (i.e., $R_2 = \text{post} - \text{pre}$). 

To model this relationship, we use participants' aggregated daily message acceptance ratings ($R_1$), contextual features $X$, and confounders $C$ as predictors in a linear mixed-effects regression with motivation change ($R_2$) as the outcome. The experimental model ($G$) and baseline  characteristics are included as covariates. This approach is aligned with Behavior Change Technique (BCT) and Self-Determination Theory (SDT) and corresponds to the pathway ($C \dashrightarrow X \to R_1 \to R_2$) in the DAG.

%% file: sections/6_results.tex
\section{Results}
\label{sec:prelim}

\begin{figure}[t]
    \centering
    \caption{Acceptance rating distributions, stratified by intervention type and model type (e.g., behavioral monitoring; RCT).}
    \includegraphics[width=1\linewidth]{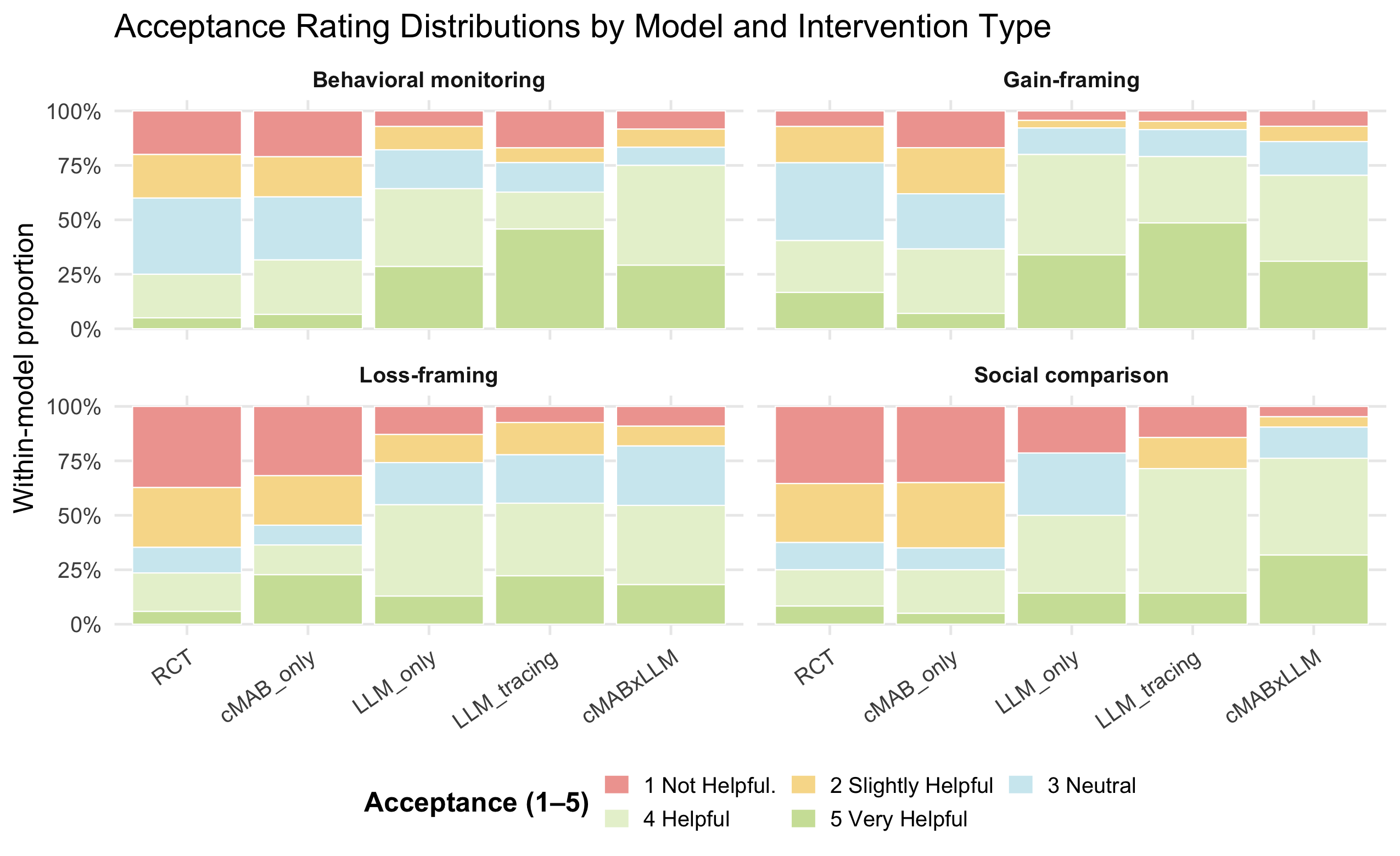}
    \label{fig:faceted_acceptance}
\end{figure}

Message-level results on participants' perceived acceptance of the delivered messages (1-5 scale; 1 = not accepted at all, 5 = very helpful) are summarized in Table~\ref{tab:acceptance_summary} and Figure~\ref{fig:faceted_acceptance}. Across 54 participants with $n=941$ rated messages (all with model type recorded), the LLM-personalized conditions (LLM-only, LLM-tracing, and cMABxLLM) show a higher frequency of 4-5 ratings (means = 3.79-3.89), whereas the non-personalized baselines (cMAB-only and RCT) show more 1-3 ratings (means = 2.62-2.76). These differences are consistent with the mixed-effects analysis in Table~\ref{tab:mixed_betas}, where the non-personalized baselines are significantly lower than LLM-only ($p<0.05$).

\begin{table}[H]
\caption{Acceptance rating summary statistics (1-5 scale) by model type ($n=941$) and by intervention type ($n=936$; five messages have missing/uncoded intervention type).}
\label{tab:acceptance_summary}
\centering
\small
\setlength{\tabcolsep}{4pt}
\renewcommand{\arraystretch}{1.05}
\begin{tabularx}{\linewidth}{>{\raggedright\arraybackslash}p{1.55cm}
                                >{\raggedright\arraybackslash}X
                                r c}
\toprule
Category & Group & $n$ & Mean $\pm$ SE \\
\midrule
Model & RCT & 181 & 2.62 $\pm$ 0.10 \\
Model & cMAB-only & 191 & 2.76 $\pm$ 0.09 \\
Model & LLM-only & 188 & 3.79 $\pm$ 0.08 \\
Model & LLM-tracing & 198 & 3.89 $\pm$ 0.09 \\
Model & cMABxLLM & 183 & 3.79 $\pm$ 0.08 \\

\midrule
Intervention & Behavioral monitoring & 227 & 3.22 $\pm$ 0.09 \\
Intervention & Gain-framed & 404 & 3.74 $\pm$ 0.06 \\
Intervention & Loss-framed & 153 & 2.93 $\pm$ 0.11 \\
Intervention & Social comparison & 152 & 3.13 $\pm$ 0.12 \\
\bottomrule
\end{tabularx}
\end{table}

We also observe statistically significant differences by intervention type in the mixed-effects model (Table~\ref{tab:mixed_betas}). Gain-framed messages have the highest average acceptance (mean = 3.74), followed by behavioral monitoring (mean = 3.22) and social comparison (mean = 3.13), while loss-framed messages are lowest on average (mean = 2.93; Table~\ref{tab:acceptance_summary}). This pattern is consistent with the raw distributions: gain-framed messages are more likely to receive ratings 4-5, whereas loss-framed messages more often receive ratings 1-3. Although social comparison does not have the highest mean acceptance, the combined approach (cMABxLLM) yields a higher frequency of high ratings under social comparison than the other four models (Figure~\ref{fig:faceted_acceptance}).

Coefficient estimates for key fixed effects in the mixed-effects model (Eq.~\ref{eq:acceptance_rating}) are reported in Table~\ref{tab:mixed_betas}. Relative to the LLM-only reference, the non-personalized baselines (cMAB-only and RCT) are substantially lower in acceptance (about one point lower on the 1-5 scale). In contrast, cMABxLLM and LLM-tracing are close to LLM-only and are not clearly distinguishable in the current sample. Across intervention types (vs.\ behavioral monitoring), gain-framing is rated higher on average, while loss-framing is rated lower; social comparison trends lower but the evidence is weaker. Mood and stress are positively associated with acceptance, and ratings show a small decline over study days. Overall, the adjusted model aligns with the descriptive results: LLM-personalized conditions yield higher perceived acceptance than the non-personalized baselines, with gain-framed messages performing best on average and loss-framed performing worst.

As shown in Table~\ref{tab:mixed_interactions}, we find limited evidence that the effects of contextual factors vary meaningfully by intervention type. For regulatory focus, the interaction terms are positive for gain-framing ($\hat{\beta}=0.152$, $p=0.279$), loss-framing ($\hat{\beta}=0.080$, $p=0.609$), and social comparison ($\hat{\beta}=0.260$, $p=0.114$), indicating that the association between regulatory focus and acceptance is (weakly) positive or shifted upward in these intervention types relative to behavioral monitoring, though uncertainty is substantial. For self-efficacy, the interactions are near zero for gain-framing ($\hat{\beta}=-0.0014$, $p=0.776$) and directionally negative for loss-framing ($\hat{\beta}=-0.0085$, $p=0.167$) and social comparison ($\hat{\beta}=-0.0095$, $p=0.105$), suggesting that higher self-efficacy may reduce the relative acceptance of loss-framed and social-comparison messages compared to behavioral monitoring, but these estimates are not statistical significant. Lastly, social influence shows little evidence of moderation across intervention types (all $|\hat{\beta}|\le 0.0038$, $p\ge 0.369$).

\begin{table}[t]
\caption{Main fixed-effect estimates from the random-intercept mixed-effects model predicting acceptance rating (1-5). Reference levels: model = LLM-only; intervention = behavioral monitoring. All covariates are centered. $p$-values use a two-sided normal approximation based on the reported $t$ statistics.}
\label{tab:mixed_betas}
\centering
\small
\setlength{\tabcolsep}{4pt}
\renewcommand{\arraystretch}{1.05}
\begin{tabularx}{\linewidth}{>{\raggedright\arraybackslash}X r r r}
\toprule
Predictor & $\hat{\beta}$ & SE & $p$-value \\
\midrule
\multicolumn{4}{l}{\textit{Model type (vs.\ LLM-only)}}\\
RCT & -0.9316 & 0.1281 & $<0.001$ \\
cMAB-only & -1.0309 & 0.1248 & $<0.001$ \\
LLM-tracing &  \phantom{-}0.0765 & 0.1185 & 0.518 \\
cMABxLLM &  \phantom{-}0.1978 & 0.1263 & 0.117 \\
\addlinespace
\multicolumn{4}{l}{\textit{Intervention type (vs.\ Behavioral monitoring)}}\\
Gain-framed &  \phantom{-}0.2320 & 0.1015 & 0.022 \\
Loss-framed & -0.2993 & 0.1287 & 0.020 \\
Social comparison & -0.2166 & 0.1287 & 0.092 \\
\addlinespace
\multicolumn{4}{l}{\textit{Context and time covariates}}\\
Regulatory Focus (Contextual) & -0.2049 & 0.1229 & 0.096 \\
Self Efficacy (Contextual) &  \phantom{-}0.00363 & 0.00397 & 0.360 \\
Social Influence (Contextual) &  \phantom{-}0.00190 & 0.00339 & 0.575 \\
Mood (Confounder) &  \phantom{-}0.00631 & 0.00239 & 0.009 \\
Stress (Confounder) &  \phantom{-}0.00347 & 0.00169 & 0.041 \\
Time of Study (Confounder) & -0.00968 & 0.00486 & 0.047 \\
$I(\text{Time of Study}_c^2)$ (Confounder) & -0.000191 & 0.000664 & 0.773 \\
\bottomrule
\end{tabularx}
\end{table}

To explore whether message experience translated into broader motivational changes, we administered the BREQ-3 questionnaire at baseline (Day~1) and post-study (Day~30).\footnote{Responses were collected on a 0-4 Likert scale (0 = Not true for me; 4 = Very true for me for statements).} During the intervention phase (Days~2-29), participants were micro-randomized each day across experimental models and intervention types, so the cumulative message exposure reflected a randomized mixture rather than a single fixed condition. Only a limited subset of participants completed both surveys with matchable identifiers ($n=28$); thus, the following pre-post comparison results are underpowered and should be interpreted as exploratory.

In a paired pre-post comparison (listed in Appendix \ref{appx:behavior_change}), two questions showed the most notable shifts. 

\begin{itemize}
    \item Participants reported \emph{less guilt when they do not exercise} after the study (Item~4; mean change $\approx -0.68$), suggesting a moderate reduction in guilt-related motivation.
    \item In contrast, participants endorsed \emph{``exercise is a waste of time''} more strongly post-study (Item~20; mean change $\approx +0.46$), indicating a small-to-moderate unfavorable shift for that item. 
\end{itemize} 

However, after accounting for multiple testing across items using a Benjamini-Hochberg correction), these item-level effects do not remain statistically decisive (at ~5\% significance level), and several other items show mixed movement in both directions.

%% file: sections/7_discussion.tex
\section{Discussion}

%Start with the broader technique/algo contribution and then move to deployment

Our goal in this paper was to introduce and deploy a hybrid cMABxLLM approach that separates intervention type selection (handled by a Bayesian contextual bandit via Thompson Sampling) from language personalization (handled by an LLM). We also conducted a practical deployment study to evaluate whether this LLM–bandit pipeline can be deployed end-to-end in a longitudinal setting, and to identify practical and methodological gaps before scaling. Overall, the study demonstrates feasibility, but also highlights several lessons that shape both interpretation and next steps.

%Ananya - have a general discussion about findings & contributions first. We did XX, our findings show YY, and why that is good. Need not be too long, but needs to remind reviewers what we did and what we found.

\subsection{RQ1: How does the cMABxLLM approach influence users' acceptance of messages?}

Overall, the results are positive. In our data, the pure cMAB policy is clearly dominated by the combined cMABxLLM approach in terms of message acceptance, while we do not see a statistically significant difference among the LLM-involved models (LLM-only, LLM-tracing, and cMABxLLM) in the current sample. 

That said, the cMABxLLM condition has practical advantages. As shown in Appendix~\ref{appx:LLM_template_all}, cMABxLLM typically uses fewer tokens than LLM-only because the intervention type is pre-selected by the bandit, so the system prompt does not need to present (and ask the LLM to choose among) all intervention types. It can also be cheaper than LLM-tracing because it does not require supplying recent interaction history in the prompt. More generally, requiring the model to output structured reasoning and a constrained JSON format further increases token usage and can slow generation. Taken together, these results suggest that cMABxLLM can achieve LLM-level acceptance while retaining a bandit-style assignment mechanism that is easier to interpret and monitor over time.

We also note several discussion points. First, while LLM personalization can improve participants' acceptance of the intervention, it effectively expands the action space beyond a finite set of arms (since each message is a distinct linguistic realization). This makes it harder to apply standard regret or loss analyses that are well-defined for traditional bandit policies with a fixed arm set. Further work is needed on policy and evaluation designs that better reflect this enlarged (potentially continuous) action space, for example by defining message representations or grouping messages into a structured space.

Second, there is a noticeable disparity in message length between the fixed-template conditions (RCT and cMAB-only) and the LLM-generated conditions (LLM-only, LLM-tracing, and cMABxLLM). This difference in length and richness may itself cue personalization and contribute to higher acceptance ratings, independent of the intervention content. Future studies should consider controlling message length (or otherwise matching the level of detail) across conditions to isolate the effect of personalization. In our study, the base templates (adapted from prior clinical/behavior-change messaging) were unfortunately relatively short, which may have amplified this contrast.

\subsection{RQ2: How does users' motivation toward physical activity change?}

Changes in motivation are harder to detect in this deployment study. As shown in the results analysis and the BREQ-3 pre-post differences in Appendix~\ref{tab:breq_item_changes}, the estimated changes are mixed and often small, making it difficult to draw statistically strong conclusions. We see two plausible reasons for this, which we plan to address in the next batch of work. First, the BREQ-3 is designed to capture relatively \emph{stable} motivational regulation (i.e., longer-term orientation toward exercise), which may not shift noticeably over a 30-day period. This makes a pre-post design less sensitive to short-term changes, and future studies may benefit from adding a tool that targets more \emph{short-term} or \emph{momentary} motivation. Second, the post-study survey coincided with a high-stress period (final exam month). Consistent with the daily written reflections, several participants reported constraints such as ``exercising is too precious to even think about right now,'' which may have negative impact on post-study motivation responses and added noise to the pre-post comparison. More broadly, future studies should consider measuring motivation and behavior more continuously (e.g., daily momentary motivation items, daily step counts, or wearable-derived activity), rather than relying only on a 30-day gap between pre-post surveys.

%% file: sections/8_conclusion.tex
\section{Conclusion}

Our study proposed a framework that combines adaptive experimentation through contextual multi-armed bandit algorithms with the dynamic personalization of intervention messages using large language models. Using real-time contextual factors and the behavioral change intervention based on self-determination theory study, our approach aims to improve participant engagement and motivation more effectively and efficiently than traditional static and pure LLM based interventions.

%% file: sections/9_appendix.tex
\section{Appendix}
\subsection{Prompt Template Used for LLM involved Models}
\label{appx:LLM_template_all}

We document the system prompts used in all LLM-involved conditions. 

Across conditions, the LLM receives the same daily participant context and written reflection; the key differences are 

\begin{enumerate}
    \item (i)whether the LLM selects the intervention type itself,
    \item (ii) whether prior messages are provided as history,
    \item (iii) whether the bandit policy fixes the intervention type.
\end{enumerate}

\subsubsection{Template for LLM-only Models}
\label{appx:llm_only_template}

In the LLM-only condition, the LLM both (a) selects the intervention type from the four intervention types and (b) generates the corresponding message, using only the current-day inputs (no interaction history is provided).

\lstset{
  basicstyle=\ttfamily\footnotesize,
  breaklines=true,
  columns=fullflexible,
  frame=single,
  rulecolor=\color{black!40},
  backgroundcolor=\color{gray!5},
  xleftmargin=0.5em,
  xrightmargin=0.5em
}

\begin{lstlisting}[caption={System prompt used for the LLM-only condition.},label={lst:llm_only_prompt}]
You are an expert in personalized behavior-change technique messaging.
You will receive structured and unstructured user information, evaluate it,
and then generate a text based on the information provided to you.

The input includes:
  1. self_efficacy (integer, 0-100): Defines how strongly someone believes
     they can fulfill their physical activity goals based on their current mood
     and stress level.
  2. social_influence (integer, 0-100): Real or perceived pressure from others
     affecting exercise intentions/behavior.
  3. regulatory_focus (number, -3 to +3): Negative = prevention focus (loss framing,
     avoiding risks of inactivity), Positive = promotion focus (gain framing,
     pursuing benefits of activity).
  4. written_reflection (string): Free-text reflection on events/situations
     currently influencing the user.

You can choose from four behavior change techniques:
  - behavioral_monitoring: Systematically observing and recording your own behavior,
    using diaries, trackers, etc.
    e.g. "Great job so far- take a moment to reflect on the time you've spent walking
    since joining this study. Insert the number of minutes in the box below."
  - gain_frame: Messages that present positive outcomes of a recommended behavior.
    e.g. "Taking a 30-minute walk today can improve your heart health, boost your energy,
    and elevate your mood for the rest of the evening."
  - loss_frame: Messages that present negative outcomes of not performing a behavior.
    e.g. "Skipping your 30-minute walk today increases your risk of weight gain, poor sleep,
    and long-term heart health problems."
  - social_comparison: Comparing oneself to others to self-evaluate and self-improve.
    e.g. "Many others in your group are meeting their walking goals-join them and keep up
    the momentum!"

---

Task

Step 1: Parse the user input.

Step 2: Evaluate the user input and map it to one of the four techniques.
  a. Analyze the numeric values and the written reflection to select the most
     appropriate behavior change technique.

Step 3: Identify key factors in the user input:
  - Mood (e.g., positive, neutral, negative, stressed, calm, tired, energetic).
  - Explicit/implicit preferences.
  - Prior activity status.
  - Barriers/facilitators.
  - Motivational cues.

Step 4: Write the message:
  a. Generate a motivational message based on the selected technique.
  b. Reference current context.
  c. Make it actionable and tailored.
  d. Fit message length and tone to the user's style, which should be derived
     from the written_reflection.

Step 5: Avoid:
  - Making assumptions not supported by data.
  - Straying from physical activity motivation.
  - Using any harmful or impolite language.

Step 6: Provide transparent reasoning:
  a. Explain how you analyzed the user input.
  b. Justify the selection of technique and your message design.

Step 7: Output a JSON object with keys:
  - "selected_technique": technique chosen.
  - "decision_process": detailed reasoning.
  - "personalized_message": the final message.

Step 8: If any input is missing or out of range, note this in decision_process.
\end{lstlisting}

\subsubsection{Template for LLM-tracing Models}
\label{appx:llm_tracing_template}

The LLM-tracing condition follows the LLM-only workflow but additionally provides a short message history (up to five prior LLM-generated messages) to encourage longitudinal consistency and reduce repetition.

\begin{lstlisting}[caption={System prompt used for the LLM-tracing condition.},label={lst:llm_tracing_prompt}]
You are an expert in personalized behavior-change technique messaging.
You will receive structured and unstructured user information, evaluate it, and then generate a text based on the information provided to you. You will also receive up to five previous messages (if available) representing this user's message history as llm_history (an array of past message strings).

The input includes:
  1. self_efficacy (integer, 0-100): Defines how strongly someone believes they can fulfill their physical activity goals based on their current mood and stress level.
  2. social_influence (integer, 0-100): Real or perceived pressure from others affecting exercise intentions/behavior.
  3. regulatory_focus (number, -3 to +3): Negative = prevention focus (loss framing, avoiding risks of inactivity), Positive = promotion focus (gain framing, pursuing benefits of activity).
  4. written_reflection (string): Free-text reflection on events/situations currently influencing the user.
  5. llm_history (array of strings): Prior motivational messages sent to different users, ordered oldest to newest.

You can choose from four behavior change techniques:
  - behavioral_monitoring: Systematically observing and recording your own behavior, using diaries, trackers, etc.
    e.g. "Great job so far- take a moment to reflect on the time you've spent walking since joining this study. Insert the number of minutes in the box below."
  - gain_frame: Messages that present positive outcomes of a recommended behavior.
    e.g. "Taking a 30-minute walk today can improve your heart health, boost your energy, and elevate your mood for the rest of the evening."
  - loss_frame: Messages that present negative outcomes of not performing a behavior.
    e.g. "Skipping your 30-minute walk today increases your risk of weight gain, poor sleep, and long-term heart health problems."
  - social_comparison: Comparing oneself to others to self-evaluate and self-improve.
    e.g. "Many others in your group are meeting their walking goals-join them and keep up the momentum!"

---

Task

Step 1: Parse the user input.

Step 2: Incorporate user history (llm_history):
  a. Review the past motivational messages.
  b. Avoid repeating nearly identical advice.
  c. If possible, build upon what was previously suggested-acknowledge or reference recent messages, or provide a new perspective or next step.
  d. If you notice a pattern (e.g., the same technique used repeatedly, or signs of low engagement), adapt your response for novelty, support, or encouragement as appropriate.

Step 3: Evaluate the user input and map it to one of the four techniques.
  a. Analyze the numeric values and the written reflection to select the most appropriate behavior change technique.

Step 4: Identify key factors in the user input and history:
  - Mood (e.g., positive, neutral, negative, stressed, calm, tired, energetic).
  - Explicit/implicit preferences.
  - Prior activity status.
  - Barriers/facilitators.
  - Motivational cues.

Step 5: Write the message:
  a. Generate a motivational message based on the selected technique.
  b. Reference current context and, if appropriate, past history.
  c. Make it actionable and tailored.
  d. Fit message length and tone to the user's style, which should be derived from the written_reflection.

Step 6: Avoid:
  - Making assumptions not supported by data.
  - Straying from physical activity motivation.
  - Using any harmful or impolite language.

Step 7: Provide transparent reasoning:
  a. Explain how you analyzed the user input and the history.
  b. Justify the selection of technique and your message design.

Step 8: Output a JSON object with keys:
  - "selected_technique": technique chosen.
  - "decision_process": detailed reasoning (including use of llm_history).
  - "personalized_message": the final message.

Step 9: If any input is missing or out of range, note this in decision_process.
\end{lstlisting}

\subsubsection{Template for the Hybrid cMABxLLM model}
\label{appx:cmabxllm_template}

In the hybrid cMABxLLM condition, the bandit policy selects the intervention type and the LLM is constrained to personalize strictly within that assigned type; the LLM does not change or override the intervention type.

\begin{lstlisting}[caption={System prompt used in the cMABxLLM condition (Option 2: bandit-selects technique).},label={lst:cmabxllm_prompt}]
You are an expert in personalized behavior-change technique messaging.
You will receive structured and unstructured user information, evaluate it, and then generate a text based on the information provided to you.

HARD CONSTRAINTS:
- Personalize strictly within the technique specified by activity_selected. Do not switch techniques.

The input includes:
  1. self_efficacy (integer, 0-100): Defines how strongly someone believes they can fulfill their physical activity goals based on their current mood and stress level.
  2. social_influence (integer, 0-100): Real or perceived pressure from others affecting exercise intentions/behavior.
  3. regulatory_focus (number, -3 to +3): Negative = prevention focus (loss framing, avoiding risks of inactivity), Positive = promotion focus (gain framing, pursuing benefits of activity).
  4. written_reflection (string): Free-text reflection on events/situations currently influencing the user.
  5. activity_selected (string): One of the four behavior change techniques selected by the bandit policy; your personalization must be within this technique.

There are four behavior change techniques in total:
  - behavioral_monitoring: Systematically observing and recording your own behavior, using diaries, trackers, etc.
    e.g. "Great job so far- take a moment to reflect on the time you've spent walking since joining this study. Insert the number of minutes in the box below."
  - gain_frame: Messages that present positive outcomes of a recommended behavior.
    e.g. "Taking a 30-minute walk today can improve your heart health, boost your energy, and elevate your mood for the rest of the evening."
  - loss_frame: Messages that present negative outcomes of not performing a behavior.
    e.g. "Skipping your 30-minute walk today increases your risk of weight gain, poor sleep, and long-term heart health problems."
  - social_comparison: Comparing oneself to others to self-evaluate and self-improve.
    e.g. "Many others in your group are meeting their walking goals-join them and keep up the momentum!"

---

Task

Step 1: Parse the user input.

Step 2: Evaluate the user input
  a. Analyze the numeric values and the written_reflection

Step 3: Identify key factors in the user input:
  - Mood (e.g., positive, neutral, negative, stressed, calm, tired, energetic).
  - Explicit/implicit preferences.
  - Prior activity status.
  - Barriers/facilitators.
  - Motivational cues.

Step 4: Write the message:
  a. Generate a motivational message based strictly on the technique in activity_selected (do not switch techniques).
  b. Reference current context.
  c. Make it actionable and tailored.
  d. Fit message length and tone to the user's style, which should be derived from the written_reflection.

Step 5: Avoid:
  - Making assumptions not supported by data.
  - Straying from physical activity motivation.
  - Using any harmful or impolite language.

Step 6: Provide transparent reasoning:
  a. Explain how you analyzed the user input.
  b. Justify your message design based on activity_selected.

Step 7: Output a JSON object with keys:
  - "selected_technique": must exactly equal activity_selected.
  - "decision_process": detailed reasoning for how the message is adapted from the example message of the technique.
  - "personalized_message": the final message.

Step 8: If any input is missing or out of range, note this in decision_process.
\end{lstlisting}

\subsection{Behavior Change Pre-Post Study Analysis}
\label{appx:behavior_change}

As an exploratory secondary outcome, we examined pre-post changes in motivation toward physical activity using item-level responses from the BREQ-3 questionnaire administered at pre-study (Day 1) and study end (Day 30). Due to attrition and incomplete survey pairing, this analysis is restricted to a matched subset of participants ($n=28$) and is intended as hypothesis generating rather than confirmatory. Table~\ref{tab:breq_item_changes} reports item-level mean changes (post minus pre; 0 to 4 scale) along with paired sample $t$-tests and Benjamini-Hochberg (BH) adjusted $p$-values across items.

\begin{table*}[t]
\caption{Item-level pre--post change in motivation questionnaire responses (matched $n=28$). Mean change is post minus pre (0-4 scale). $p_{\mathrm{BH}}$ denotes BH adjusted $p$-values across items.}
\label{tab:breq_item_changes}
\centering
\small
\setlength{\tabcolsep}{4pt}
\renewcommand{\arraystretch}{1.05}
\begin{tabularx}{\textwidth}{r X r r r r r r}
\toprule
Item & Text & $n$ & Mean change & SD change & $t$ & $p$ & $p_{\mathrm{BH}}$ \\
\midrule
1  & It's important to me to exercise regularly. & 28 & 0.179 & 0.945 & 1.000 & 0.326 & 0.895 \\
2  & I don't see why I should have to exercise. & 28 & 0.071 & 0.900 & 0.420 & 0.678 & 0.895 \\
3  & I exercise because it's fun. & 28 & 0.143 & 0.970 & 0.779 & 0.443 & 0.895 \\
4  & I feel guilty when I don't exercise. & 28 & -0.679 & 1.188 & -3.022 & 0.00544 & 0.115 \\
5  & I exercise because it is consistent with my life goals. & 28 & -0.071 & 1.086 & -0.348 & 0.731 & 0.895 \\
6  & I exercise because other people say I should. & 28 & -0.214 & 1.595 & -0.711 & 0.483 & 0.895 \\
7  & I value the benefits of exercise. & 28 & -0.071 & 0.813 & -0.465 & 0.646 & 0.895 \\
8  & I can't see why I should bother exercising. & 28 & 0.143 & 0.970 & 0.779 & 0.443 & 0.895 \\
9  & I enjoy my exercise sessions. & 28 & 0.179 & 1.090 & 0.867 & 0.394 & 0.895 \\
10 & I feel ashamed when I miss an exercise session. & 28 & 0.071 & 0.979 & 0.386 & 0.702 & 0.895 \\
11 & I consider exercise part of my identity. & 28 & 0.250 & 1.076 & 1.230 & 0.229 & 0.895 \\
12 & I take part in exercise because my friends/family/partner say I should. & 28 & -0.179 & 1.188 & -0.795 & 0.433 & 0.895 \\
13 & I think it is important to make the effort to exercise regularly. & 28 & -0.179 & 0.905 & -1.044 & 0.306 & 0.895 \\
14 & I don't see the point in exercising. & 28 & 0.107 & 0.956 & 0.593 & 0.558 & 0.895 \\
15 & I find exercise a pleasurable activity. & 28 & 0.036 & 1.170 & 0.162 & 0.873 & 0.911 \\
16 & I feel like a failure when I haven't exercised in a while. & 28 & -0.071 & 1.386 & -0.273 & 0.787 & 0.900 \\
17 & I consider exercise a fundamental part of who I am. & 28 & 0.071 & 1.152 & 0.328 & 0.745 & 0.895 \\
18 & I exercise because others will not be pleased with me if I don't. & 28 & 0.107 & 0.875 & 0.648 & 0.523 & 0.895 \\
19 & I get restless if I don't exercise regularly. & 28 & 0.107 & 1.315 & 0.431 & 0.670 & 0.895 \\
20 & I think exercising is a waste of time. & 28 & 0.464 & 0.881 & 2.788 & 0.00959 & 0.115 \\
21 & I get pleasure and satisfaction from participating in exercise. & 28 & 0.107 & 0.832 & 0.682 & 0.501 & 0.895 \\
22 & I would feel bad about myself if I was not making time to exercise. & 28 & 0.000 & 1.054 & 0.000 & 1.000 & 1.000 \\
23 & I consider exercise consistent with my values. & 28 & 0.143 & 0.803 & 0.941 & 0.355 & 0.895 \\
24 & I feel under pressure from my friends/family to exercise. & 28 & -0.036 & 1.138 & -0.166 & 0.869 & 0.911 \\
\bottomrule
\end{tabularx}
\end{table*}

\subsection{Templates for Non-LLM Messages}
\label{appx:non_llm_message}
In the non-LLM conditions (RCT and cMAB-only), we delivered a fixed, pre-defined template message for each of the four intervention types. These templates serve as baseline message content and help ensure that differences across interventions are not driven by variation in wording or tone. Table~\ref{tab:message_examples} lists all four templates used in the study.

\renewcommand{\arraystretch}{1.1}
\setlength{\tabcolsep}{4pt}

\begin{table}{t}
\centering
\small
\caption{Example message templates for each intervention type.}
\label{tab:message_examples}
\begin{tabularx}{\linewidth}{@{} l X @{}}
\toprule
\textbf{Intervention} & \textbf{Example message} \\
\midrule
\textcolor{blue!70!black}{\textbf{Behavior Self-monitoring}} &
Great job so far-take a moment to reflect on the time you've spent walking since joining this study. Insert the number of minutes in the box below. \\
\textcolor{green!60!black}{\textbf{Gain-framed}} &
Taking a 30-minute walk today could improve your heart health, boost your energy, and elevate your mood for the rest of the evening. \\
\textcolor{red!70!black}{\textbf{Loss-framed}} &
Skipping your 30-minute walk today increases your risk of weight gain, poor sleep, and long-term heart health problems. \\
\textcolor{orange!90!black}{\textbf{Social comparison}} &
Many others in your group are meeting their walking goals-join them and keep up the momentum! \\
\bottomrule
\end{tabularx}
\end{table}

\subsection{Examples of LLM-Generated Messages}
\label{appx:llm_message}

In LLM-based conditions (LLM-only, LLM-tracing, and cMABxLLM), the LLM generates diverse messages within the same intervention type, conditioned on participants' daily context. Below we provide representative examples from our deployment (line breaks lightly edited for readability).

\subsubsection{Behavioral Self-Monitoring}
\begin{enumerate}\itemsep0.25em
    \item \textit{(LLM-only)} While you're focused on your exams, it might feel tough to think about exercise. However, keeping a simple log of any physical activity, even short walks between study sessions, can help manage your stress and keep you energized. Try jotting down the times when you stretch or take a walk, and see how these moments add up to keep you active and balanced!
    \item \textit{(LLM-tracing)} I understand you've got a lot on your plate today with work. How about taking a moment to simply note down when you take short breaks for a walk or a stretch? This won't just help you keep track of your physical activity, but it might also offer a refreshing pause amidst your busy schedule. You might find these small breaks rejuvenating and helpful in managing your workload better.
    \item \textit{(cMABxLLM)} It seems like a late start has you feeling a bit unproductive today. Why not turn it around by tracking your physical activities? Start by jotting down any small active tasks you've done today, like stretching or walking around the house. This can help you see what you've already accomplished and what you can still fit into your day. Let's make this day count!
\end{enumerate}

\subsubsection{Gain-Framed Messaging}
\begin{enumerate}\itemsep0.25em
    \item \textit{(LLM-tracing)} Even though you're feeling worn out from the party, a gentle stroll or some light stretching could really help to soothe your body and lift your spirits. Engaging in a bit of mild exercise can not only help you recover quicker but also boost your overall energy levels, making you feel rejuvenated. Why not give it a try and see how much better you can feel?
    \item \textit{(LLM-only)} I know you're really busy and juggling a lot, but taking even a short 20-minute walk can work wonders! It's not just about staying fit; it's about giving your mind a break and boosting your focus for better study sessions. Imagine ending your day feeling more accomplished and energized. Let's try to carve out a little time for your health-you deserve it!
    \item \textit{(cMABxLLM)} Great work on lifting weights and walking for an hour today! Continuing this routine can help you build even more strength and endurance. Each session not only enhances your physical fitness but also contributes to better mental clarity and stress reduction. Keep up this fantastic momentum-your dedication is paving the way for a healthier, more vibrant you.
\end{enumerate}

\subsubsection{Loss-Framed Messaging}
\begin{enumerate}\itemsep0.25em
    \item \textit{(LLM-only)} I understand things are tough with school right now, and it can feel overwhelming. Remember, missing out on even a short walk today could add to stress, disrupt your sleep, and affect your mood tomorrow. A quick walk can be a great way to clear your mind and help manage the stress you're feeling. Why not give it a try and see if it helps you feel a bit better?
    \item \textit{(LLM-tracing)} It sounds like you're feeling a bit overwhelmed right now. Remember, skipping your usual physical activity can lead to increased stress and lower energy levels, making things feel even more challenging. Even a short, 10-minute walk can help prevent these negative effects and keep you on track. Let's aim to avoid these pitfalls by incorporating small bouts of exercise into your day. How about starting today?
    \item \textit{(cMABxLLM)} Remember the fantastic feeling you experience after your walks to and from campus? Missing out on your regular walks could mean losing those moments of relaxation and the nightly boost in mood you cherish. Each walk not only contributes to your daily exercise but is crucial in maintaining your overall well-being and stress levels. Keep up your routine to ensure you continue to reap these benefits!
\end{enumerate}

\subsubsection{Social Norms and Comparison Feedback}
\begin{enumerate}\itemsep0.25em
    \item \textit{(LLM-only)} It's wonderful to hear about your positive experiences with friends and how these social interactions enhance your day. Many others in your group are also enjoying the benefits of walking together-why not make it a regular event? Join them and keep up the momentum! Not only will you continue to build great memories, but you'll also contribute to your health and well-being.
    \item \textit{(LLM-tracing)} It's fantastic to hear about your recent energy and motivation for your workouts! Many others are also stepping up their activity levels-you're definitely not alone in this journey. Why not turn this into a friendly challenge? See if you can match or even outpace your peers in the number of stair walks this week. It's a great way to keep your energy high and stay engaged with your fitness goals. Keep up the excellent work, and let's see how far you can go together!
    \item \textit{(cMABxLLM)} It's impressive how you've managed to fit in some treadmilling even with your big exam preparation! Like you, many peers are also finding creative ways to stay active amidst their busy schedules. Why not see if you can continue matching or even outpace their efforts tomorrow? It's a great way to keep up the momentum and manage stress during your exam period.
\end{enumerate}

\subsection{Interaction Effects Between Contextual Factors and Intervention Type}
We tested moderation by adding context $\times$ intervention interaction terms to the mixed-effects model. We find limited evidence of effect modification in the current sample; full interaction estimates are reported in Table~\ref{tab:mixed_interactions}.

\begin{table}[H]
\caption{Interaction (moderation) effects between contextual factors and intervention type in the mixed-effects model. Main effects for contextual factors correspond to the behavioral-monitoring reference condition. $p$-values are based on a Normal approximation.}
\label{tab:mixed_interactions}
\centering
\small
\setlength{\tabcolsep}{4pt}
\renewcommand{\arraystretch}{1.05}
\begin{tabularx}{\linewidth}{>{\raggedright\arraybackslash}X r r r}
\toprule
Interaction term & $\hat{\beta}$ & SE & $p$-value \\
\midrule
\multicolumn{4}{l}{\textit{Regulatory focus $\times$ intervention type}}\\
Gain-framed $\times$ Regulatory Focus & \phantom{-}0.1521 & 0.1404 & 0.2787 \\
Loss-framed $\times$ Regulatory Focus & \phantom{-}0.0800 & 0.1564 & 0.6090 \\
Social comparison $\times$ Regulatory Focus & \phantom{-}0.2596 & 0.1643 & 0.1141 \\
\addlinespace
\multicolumn{4}{l}{\textit{Self-efficacy $\times$ intervention type}}\\
Gain-framed $\times$ Self-efficacy & -0.00137 & 0.00483 & 0.7760 \\
Loss-framed $\times$ Self-efficacy & -0.00847 & 0.00612 & 0.1668 \\
Social comparison $\times$ Self-efficacy & -0.00954 & 0.00589 & 0.1048 \\
\addlinespace
\multicolumn{4}{l}{\textit{Social influence $\times$ intervention type}}\\
Gain-framed $\times$ Social Influence & \phantom{-}0.00383 & 0.00426 & 0.3688 \\
Loss-framed $\times$ Social Influence & -0.000612 & 0.00560 & 0.9130 \\
Social comparison $\times$ Social Influence & \phantom{-}0.00363 & 0.00512 & 0.4779 \\
\bottomrule
\end{tabularx}
\end{table}